\documentclass[11pt,letterpaper]{article}

\usepackage[T1]{fontenc}
\usepackage{newtxtext}
\usepackage[margin=1in,headheight=22pt]{geometry}
\usepackage{microtype}
\usepackage{xcolor}
\definecolor{accent}{HTML}{496778}
\definecolor{ink}{HTML}{292824}
\definecolor{muted}{HTML}{77746E}
\definecolor{rule}{HTML}{D9D5CE}
\definecolor{abstractbg}{HTML}{F2F5F6}
\usepackage{amsmath,amssymb}
\usepackage{graphicx}
\usepackage{tikz}
\usepackage{booktabs}
\usepackage{caption}
\usepackage{siunitx}
\usepackage{multirow}
\usepackage[numbers,sort&compress]{natbib}
\usepackage{titlesec}
\usepackage{fancyhdr}
\usepackage[skins,breakable]{tcolorbox}
\usepackage[hidelinks]{hyperref}
\usepackage{cleveref}
\usepackage{sty/custom}
\hypersetup{colorlinks=true,linkcolor=accent,citecolor=accent,urlcolor=accent}

\titleformat{\section}
  {\Large\bfseries\color{ink}}{\color{ink}\thesection}{0.65em}{}
\titleformat{\subsection}
  {\large\bfseries\color{ink}}{\color{ink}\thesubsection}{0.65em}{}
\titleformat{\subsubsection}
  {\normalsize\bfseries}{\thesubsubsection}{0.65em}{}
\titlespacing*{\section}{0pt}{2.2ex plus 0.5ex minus 0.2ex}{0.8ex}
\titlespacing*{\subsection}{0pt}{1.8ex plus 0.4ex minus 0.2ex}{0.5ex}

\renewcommand{\headrulewidth}{0.35pt}
\renewcommand{\headrule}{\hbox to\headwidth{\color{rule}\leaders\hrule height \headrulewidth\hfill}}

\newtcolorbox{abstractbox}{
  enhanced,
  breakable,
  colback=abstractbg,
  colframe=abstractbg,
  boxrule=0pt,
  borderline west={2pt}{0pt}{accent},
  sharp corners,
  left=11pt,
  right=11pt,
  top=9pt,
  bottom=9pt,
  before skip=0.3em,
  after skip=0.6em
}
\renewenvironment{abstract}{%
  \begin{abstractbox}
  \setlength{\parindent}{0pt}%
  {\small\bfseries\color{accent}\MakeUppercase{Abstract}}\par\smallskip
  \small
}{%
  \end{abstractbox}
}

\newlength{\preprintlabelwidth}
\title{%
  \begin{minipage}{\textwidth}\raggedright
  \setlength{\parskip}{0pt}%
  \settowidth{\preprintlabelwidth}{\scriptsize PREPRINT}%
  \noindent{\color{accent}\vbox{%
    \hbox{\scriptsize PREPRINT}%
    \kern 2pt%
    \hrule width \preprintlabelwidth height 0.4pt%
  }}\par\vspace{1em}%
  {\LARGE\bfseries\color{ink} Map-Distribution Mixture-of-Experts for Multi-Agent Pathfinding}%
  \end{minipage}%
}
\author{%
  \begin{minipage}{\textwidth}\raggedright
  {\normalsize\bfseries\color{ink} Fumimaro Suzuki\textsuperscript{1,2}, Keisuke Okumura\textsuperscript{2}}\\[0.15em]
  \small\textcolor{muted}{\textsuperscript{1}University of Tsukuba, Tsukuba, Japan}\\[-0.2em]
  \small\textcolor{muted}{\textsuperscript{2}National Institute of Advanced Industrial Science and Technology (AIST), Tokyo, Japan}\\[-0.2em]
  \small\href{mailto:s2312431@u.tsukuba.ac.jp}{s2312431@u.tsukuba.ac.jp},
  \href{mailto:okumura.k@aist.go.jp}{okumura.k@aist.go.jp}
  \end{minipage}%
}
\date{}

\makeatletter
\renewcommand{\maketitle}{%
  \thispagestyle{plain}%
  \vspace*{0.4em}%
  \noindent\@title\par
  \vspace{0.3em}%
  \noindent\@author\par
  \vspace{0.2em}%
}
\makeatother

\begin{document}
\maketitle
\begin{abstract}
Designing capable neural policies for multi-agent pathfinding (MAPF), which map local observations to action distributions, represents an important challenge in multi-agent learning, with practical relevance to scalable collective automation problems such as warehouse logistics.
In contrast to prior studies focusing primarily on individual policy design, this study explores a meta-level \emph{Mixture-of-Experts (MoE)} approach to learning-based MAPF.
Specifically, given a collection of compact yet high-performing experts specialised for different map distributions, our architecture learns a router that selectively fuses their outputs according to each agent's local situation.
We instantiate and validate this MoE approach with a commonly used graph neural network-based policy as well as a newly developed recursive action-token refinement model.
Combined with supervised and reinforcement learning fine-tuning of the experts, our experiments show that MoE generally improves upon the pretrained policy and outperforms existing leading methods.
At the same time, our findings reveal that appropriately trained, parameter-matched monolithic models still outperform MoE, highlighting a nuanced role for map-distribution MoE for MAPF.
\end{abstract}

\noindent{\small\textcolor{muted}{Code -- }\url{https://github.com/romanohu/MDMoE-MAPF}}
\par
\section{Introduction}
Multi-agent pathfinding (MAPF) has emerged as a challenging testbed for large-scale multi-agent learning, requiring dense coordination among agents, while also being closely connected to real-world applications such as logistics automation.
Among the various paradigms for integrating learning into planning, one central topic is the development of capable \emph{neural MAPF policies}, in which each agent selects actions based on local observations and interactions~\cite{alkazzi2024comprehensive}.
In fact, various policy architectures have been developed, together with efficient training pipelines based primarily on reinforcement learning (RL) and imitation learning (IL).
Recent studies have further shown that integrating neural MAPF policies with search can outperform purely search-based MAPF solvers in certain problem regimes~\cite{jiang2025deploying,jain2026graph}.

Existing architectures, however, largely rely on monolithic policies, where a single policy is expected to handle all situations encountered during execution.
In contrast, \emph{Mixture-of-Experts (MoE)} architectures~\cite{jacobs1991adaptive} have proven effective across other domains.
At its core, MoE combines multiple specialised experts through a learned mechanism that selectively uses their outputs according to the given input.
A notable example is the development of large language models (LLMs), where modern architectures increasingly adopt sparse MoE to efficiently leverage specialised experts~\cite{fedus2022switch,dai2024deepseekmoe}.
Related ideas of combining specialised skills have also been explored in robotics, enabling increasingly versatile and complex behaviours~\cite{yang2020multi,hoeller2024anymal}.

This naturally raises the question: \emph{can we exploit the power of MoE for neural MAPF policies, and if so, how?}
This work addresses this question through a systematic investigation.

{
  \begin{figure}[t]
    \centering
    \includegraphics[width=0.52\linewidth]{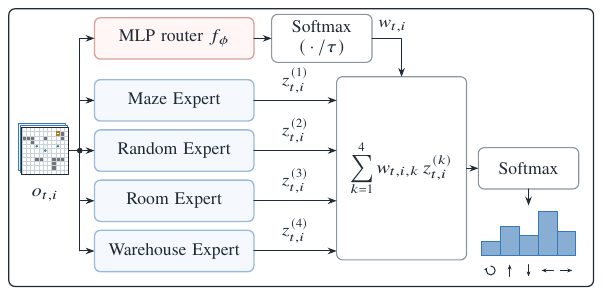}
    \captionsetup{skip=2pt}
    \caption{MoE framework.
      A learned router fuses map-distribution experts taking each agent's local observation.
    }
    \label{fig:overview}
    \label{fig:routermoe}
  \end{figure}
}

More precisely, two key considerations arise: \emph{(i)}~how to prepare the experts and \emph{(ii)}~how to combine them.
For the former, inspired by the observation in~\cite{jain2026graph} that fine-tuning a neural MAPF policy for a specific map can substantially enhance its performance, we prepare a collection of experts specialised for different map distributions.
To this end, we adopt a training strategy inspired by recent LLM development: starting from a pretrained base model, we perform supervised fine-tuning (SFT) for each target map distribution, followed by RL fine-tuning using Group Relative Policy Optimisation (GRPO)~\cite{shao2024deepseekmath}.
For the latter, we learn a routing function that takes the same observations as the experts and assigns agent-wise weights to fuse their outputs, as shown in \cref{fig:overview}.
Through this mechanism, the model can selectively leverage different experts according to the local geometric features encountered at test time.

We show that this simple recipe works effectively with two distinct neural architectures for the expert policies.
The first is a graph neural network (GNN)-based architecture called MAGAT (\emph{M}essage-\emph{A}ware \emph{G}raph \emph{At}tention Network)~\cite{li2021message,jain2026graph}, which has served as a common architectural backbone in the literature.
The second is our newly developed lightweight architecture, ReCAT (\emph{Re}cursive \emph{C}ollective \emph{A}ction \emph{T}oken Network), which recursively refines action-token representations through GNN-based token interactions within and between agents.
In particular, ReCAT-MoE outperforms existing leading methods~\cite{andreychuk2025finetuning,jain2026hmagat}, despite having fewer trainable parameters, across both in-distribution and out-of-distribution scenarios.
Meanwhile, applying the same SFT+RL recipe to parameter-matched monolithic counterparts produces remarkably strong policies, which remain competitive with or even outperform their MoE counterparts.
Together, these results demonstrate both the effectiveness and limitations of map-distribution MoE, revealing a nuanced role for this previously under-explored paradigm in neural MAPF.

The remainder of this paper is organised as follows.
We first introduce the problem formulation and basic assumptions underlying neural MAPF policies in \cref{sec:prelim}.
We then present map-wise expert training (\cref{sec:experts}), MoE training (\cref{sec:moe}), and ReCAT (\cref{sec:recursive}), followed by their empirical evaluation in \cref{sec:experiments}.
Finally, we position our study within the broader literature and conclude the paper.

\section{Preliminaries}
\label{sec:prelim}
\subsection{Problem Formulation}
We consider a commonly used MAPF formulation~\cite{stern2019def}, defined by a set of agents $A = \{1, 2, \ldots, n\}$ operating on a four-connected grid $G=(V,E)$, where each agent $i \in A$ has distinct start and goal vertices, denoted by $s_i, g_i \in V$.
At each timestep, each agent either stays at its current vertex or moves to an adjacent vertex.
Vertex and edge collisions are prohibited; that is, two agents cannot occupy the same vertex at the same timestep or traverse the same edge in opposite directions within one timestep.
The objective is to derive a collision-free path $[v_i^0 = s_i, v_i^1, \ldots, v_i^T = g_i]$ for each agent $i \in A$, where the plan length $T$ is called \emph{makespan}.
Equivalently, using a \emph{configuration} $Q \in V^n$ that represents the locations of all agents, MAPF seeks a sequence of transitionable collision-free configurations that moves every agent from its start to its goal.
The solution cost is measured by the \emph{sum-of-costs} (SoC), defined as the sum of the travel times of all agents until they reach their goals and remain there.
Following common practice, we use SoC normalised by $\sum_{i\in A}\dist(s_i,g_i)$ as cost, where $\dist(u,v)$ denotes the length of a shortest path between two vertices $u,v \in V$.

\subsection{Neural MAPF Policies}
This study regards MAPF as a collective sequential decision-making problem, where each agent $i$ receives a local observation $o_i$ from the current configuration $Q$ and selects its action according to the action distribution $\pi(o_i)$ over five possible actions, including staying in place.
Our objective is then to learn a policy $\pi$ that solves MAPF while achieving low SoC.
Such a policy is typically trained via either RL or IL from expert search-based MAPF planners.
Many policy architectures have been developed, with representative examples including PRIMAL~\cite{sartoretti2019primal}, MAPF-GPT~\cite{andreychuk2024mapfgpt,andreychuk2025finetuning}, and HMAGAT~\cite{jain2026hmagat};
see~\cite{alkazzi2024comprehensive} for a comprehensive review.
Beyond improving individual policy architectures, this study explores an orthogonal, architecture-agnostic direction: how to effectively construct experts for target domains, and fuse them through MoE.

\subsubsection{Observation}
While the proposed MoE framework itself does not assume a specific form of observation, following common practice in the literature~\cite{sartoretti2019primal,li2021message,jain2026graph}, we use an observation tensor $o_i$ centred around agent $i$'s current position $Q[i] \in V$, with shape $4\times(2R\sub{obs}+1)^2$, where $R\sub{obs} \in \mathbb{N}_{>0}$ determines the field-of-view (FOV) size.
The four channels represent:
\emph{(i)}~the presence of obstacles,
\emph{(ii)}~that of other agents,
\emph{(iii)}~a projection of the goal direction, all represented by binary values, and
\emph{(iv)}~cost-to-go values, where the value is normalised as $(\dist(v, g_i) - \dist(Q[i], g_i)) / (2R\sub{obs})$ for each location $v \in V$ within the FOV.

\subsubsection{GNN-based Policy and Communication}
To validate the efficacy of MoE, this study conducts expert fusion over two distinct neural MAPF policy architectures.
One is a newly developed architecture designed to achieve high performance with a compact policy, described in detail in \cref{sec:recursive}, while the other follows a standard GNN-based policy, MAGAT~\cite{li2021message,jain2026graph}, as described below.%
\footnote{
For ease, our implementation uses a slightly modified version of the original MAGAT architecture.
Our message-passing layers use GATv2~\cite{brody2021attentive}, while the original architecture uses a custom graph attention operation.
We continue to refer to our implementation as MAGAT, as its fundamental architecture remains unchanged.
}

Given a configuration $Q$, MAGAT assumes a communication graph $G\sub{comm}$, where each vertex corresponds to an agent and an edge $(i,j)$ exists if agents $i$ and $j$ are within a communication radius $R\sub{comm} \in \mathbb{N}^+$ of each other.
MAGAT then consists of
\emph{(i)}~a CNN-based encoder that maps the observation $o_i$ to an initial latent embedding $z_i^0$,
\emph{(ii)}~$L$ communication layers that perform GNN message passing over $G\sub{comm}$, where each layer $l$ produces $z_i^l$ from the embeddings of the previous layer, and
\emph{(iii)}~a multilayer perceptron (MLP) decoder that maps $z_i^L$ to an action distribution.
The communication layers use a three-dimensional edge feature $\omega_{ij}$, consisting of the relative position in the $x$- and $y$-directions and the Manhattan distance between the corresponding agents.

\subsubsection{Collision Shielding}
A neural MAPF policy alone outputs an action distribution, leaving a design choice in how to decode it into collision-free actions.
Earlier work stochastically samples actions from the distribution and then resolves inter-agent collisions by simply stopping one of the conflicting agents, an approach known as naive \emph{collision shielding (CS)}~\cite{li2020graph,skrynnik2025pogema}.
Meanwhile, recent studies have shown that leveraging PIBT~\cite{okumura2022priority} to sample collision-free actions substantially improves the learned policy, an approach called \emph{CS-PIBT}~\cite{veerapaneni2024improving,jiang2025deploying,veerapaneni2025work,jain2026graph,jain2026hmagat}; we mainly follow this strategy.

PIBT is a popular MAPF algorithm underlying modern scalable MAPF planners in both learning-free and learning-based settings~\cite{okumura2024lacam3,jiang2025deploying,jain2026hmagat}.
It receives a configuration and an ordered action list for each agent, called a \emph{preference}, and generates a collision-free next configuration while respecting the agent-wise preferences.
The original PIBT constructs preferences using the cost-to-go values, $\dist(\cdot,g_i)$, while CS-PIBT instead uses the learned policy for this purpose by either stochastically or deterministically sampling an ordered action list according to the assigned action probabilities.

\section{MoE over Map-Specific Experts}

A key consideration in constructing an MoE is how to prepare a diverse and complementary set of expert policies.
Two general strategies are commonly used: \emph{(i)}~training the experts first and subsequently learning a routing function to combine them, and \emph{(ii)}~jointly training the experts and router.
We adopt the first strategy, motivated by prior work~\cite{jain2026graph} showing that supervised fine-tuning of a pretrained policy on \emph{one} target map can substantially improve its performance on that map.
Following this insight, we first train a base policy and then independently fine-tune copies of it into a collection of experts, each specialised for a particular \emph{map distribution}, before learning the router that combines their outputs.
Such a training recipe has been used in MoE-based LLMs~\cite{sukhbaatar2024branch}.

Based on common benchmark categorisations~\cite{shen2023tracking,skrynnik2025pogema}, this study considers four distinct map distributions: \mapname{Maze}, \mapname{Warehouse}, \mapname{Room}, and \mapname{Random}.
We select their instance configurations while accounting for difficulty, such that they provide useful training signals for high-performing policies.
These map distributions are currently selected manually, leaving their automatic selection to future work.

\subsection{Expert Training}
\label{sec:experts}

Starting from a pretrained policy, our reproducible fine-tuning pipeline for each target map distribution, applicable across different neural architectures, consists of supervised fine-tuning (SFT) followed by RL-based fine-tuning.
\subsubsection{Pretraining and Supervised Fine-Tuning (SFT)}
We pretrain a policy to imitate teacher trajectories generated by an expert MAPF solver.
All agents share the policy parameters $\theta$, and training minimises the cross-entropy between the predicted action distributions and the teacher actions, as commonly adopted in IL-based MAPF policies.
The resulting parameters $\theta_0$ provide a common initialisation for adaptation to the target map distributions.

Next, for each target map distribution $k$, we collect additional teacher trajectories by solving instances sampled from that distribution by the expert MAPF solver.
We initialise the corresponding expert with $\theta_k=\theta_0$ and fine-tune it on these trajectories using the same cross-entropy objective.

\subsubsection{Reinforcement Learning Fine-Tuning}
\label{sec:rl-ft}

While SFT alone provides decent performance, we empirically observe that RL fine-tuning further improves each SFT expert $\pi_{k}$ using rewards from its own rollouts on the target map distribution.
Our approach consists of two key components: PIBT-based safeguarded rollouts and GRPO-based~\cite{shao2024deepseekmath} policy updates.

\paragraph{Rollout Procedure}
For each instance, we collect $B$ rollouts from the same initial configuration, ending each rollout when all agents reach their goals or the timestep limit $T_{\max}$ is reached.
At each timestep, CS-PIBT~\cite{veerapaneni2024improving} converts the policy's output distributions into a collision-free joint move.
For each agent, it samples an ordering of legal actions without replacement according to the policy probabilities, which serves as the preference ordering for PIBT.
PIBT then sequentially attempts to assign compatible moves according to these preferences, trying subsequent candidates when necessary.
For the policy update, we record the prefix of each ordering up to the last action examined;
i.e., $u_{t,i}=[v^{(1)},v^{(2)},\ldots,v^{(m)}]$ denotes the sequence of candidate vertices attempted for agent $i$ at timestep $t$ during PIBT.
This prefix is subsequently used to compute the likelihood for the policy update.

\paragraph{Reward Design}
Our reward design prioritises solving the instance, with secondary terms favouring shorter successful rollouts and a larger fraction of agents reaching their goals in unsuccessful rollouts.
More concretely, for rollout $b$, the reward is computed based on its length $T_b$ and the fraction $q_b$ of agents that have reached their goals at termination, with a hyperparameter $\eta \in \mathbb{R}^+$:
\begin{equation}
  R_b =
  \begin{cases}
    1+\eta(1-T_b/T_{\max}), & \text{if all agents reach their goals},\\
    -1+\eta q_b, & \text{otherwise}.
  \end{cases}
  \label{eq:team-score}
\end{equation}
Following GRPO, we next compute an advantage $A_b=(R_b-\bar R)/\sigma_R$, where $\bar R$ and $\sigma_R$ are the mean and standard deviation of rewards across the $B$ rollouts for the same instance.
This advantage is shared by all recorded decisions in rollout $b$.

\paragraph{Preference Likelihood}
To use these rollout-level advantages for policy optimisation, we next quantify the likelihood of each recorded preference under the policy.
For agent $i$ at timestep $t$, let $P_\theta(u_{t,i}\mid o_{t,i})$ denote the likelihood of the recorded prefix $u_{t,i}$ given observation $o_{t,i}$.
Since the preference ordering is sampled without replacement, this likelihood follows the Plackett--Luce model~\cite{plackett1975analysis}, given by the product of successive sampling probabilities, each renormalised over the remaining legal actions.
We compare this likelihood under the current parameters $\theta$ and the rollout parameters $\theta_{\mathrm{old}}$ through the importance ratio,
\begin{equation}
  r_{t,i}
  = \frac{P_\theta(u_{t,i}\mid o_{t,i})}
         {P_{\theta_{\mathrm{old}}}(u_{t,i}\mid o_{t,i})}.
  \label{eq:prefix-ratio}
\end{equation}

\paragraph{Loss Design}
We minimise a clipped policy loss~\cite{schulman2017ppo} with a Kullback--Leibler (KL) penalty relative to the frozen supervised expert $\pi_{\mathrm{ref}}$ and an entropy bonus, which prevents the policy from becoming deterministic too early, as commonly adopted in PPO training, with hyperparameters $\epsilon$, $\beta$, and $\alpha$, as follows:
\begin{equation}
  \begin{split}
    \mathcal{L}_{\mathrm{RL}}(\theta)
    ={}-\mathbb{E}\!\left[
      \min\!\left(r_{t,i}A_b,\operatorname{clip}(r_{t,i},1-\epsilon,1+\epsilon)A_b\right)
    \right]
    +\beta\mathbb{E}\!\left[
      D_{\mathrm{KL}}(\pi_{\mathrm{ref}}\Vert\pi_\theta)
    \right]
    -\alpha\mathbb{E}\!\left[\mathcal{H}(\pi_\theta)\right].
  \end{split}
  \label{eq:rl-loss}
\end{equation}
The KL divergence and entropy are evaluated at $o_{t,i}$ after restricting the action distributions to legal actions.
The expectations are taken over all recorded agent--timestep pairs $(i,t)$ across the $B$ rollouts.

{
  \begin{figure*}[t]
    \centering
    \includegraphics[width=\textwidth]{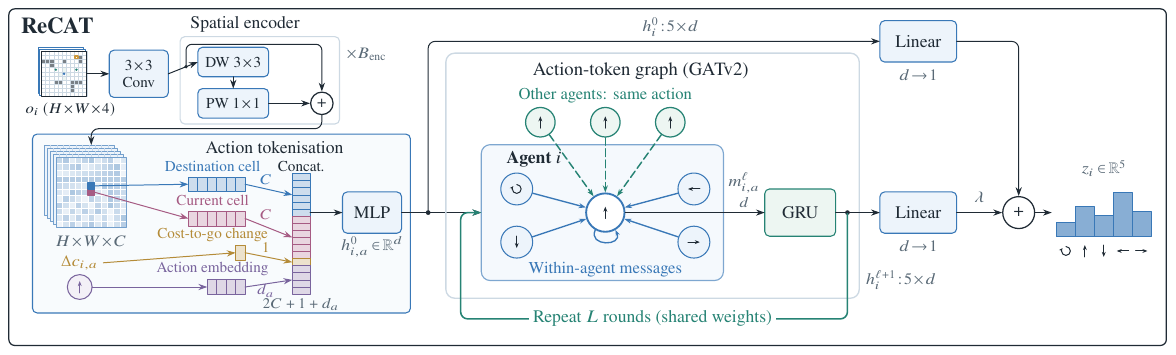}
    \caption{ReCAT architecture.
    Following the spatial encoding (left), an action token is refined through message passing from other tokens of agent $i$ and matching action tokens of neighbouring agents (right).
    Tensor shapes use $H=W=2R\sub{obs}+1$, $C$ feature channels, token dimension $d$, and action embedding dimension $d_a$.}
    \label{fig:recat}
  \end{figure*}
}

\subsection{Expert Fusion}
\label{sec:moe}

Through the training pipeline in \cref{sec:experts}, suppose that we obtain $K$ expert policies $\{\pi_{\theta_1}, \ldots, \pi_{\theta_K}\}$.
For agent $i$ at timestep $t$, expert $k$ produces a vector of five action logits $z_{t,i}^{(k)}$, which would normally be converted into an action distribution via softmax.
We instead combine these specialised policies through an observation-based MoE, as illustrated in \cref{fig:routermoe}.

\paragraph{Router Design}
To weight the expert outputs, an MLP router $f_\phi$ maps the flattened local observation $o_{t,i}$ for agent $i$ at timestep $t$ to one score per expert.
A softmax with temperature $\tau$ converts the scores into mixing weights
$w_{t,i} = \operatorname{softmax}\!\left(f_\phi(o_{t,i})/\tau\right)$.
The $k$-th component $w_{t,i,k}$ weights the logits of expert $k$, and the weighted sum of the logits is converted to the action distribution with:
\begin{equation}
  \pi_\phi(o_{t,i})
  = \operatorname{softmax}\!\left(
    \sum_{k=1}^{K} w_{t,i,k}\,z_{t,i}^{(k)}
  \right).
  \label{eq:moe-policy}
\end{equation}
This defines our MoE policy, where the contribution of each expert is determined independently for each agent and timestep from its local observation.

\paragraph{Training}
Once the experts have been trained, we freeze their parameters and optimise only the router parameters $\phi$ on the concatenation of the supervised fine-tuning datasets across all map distributions.
We train the combined policy using the same cross-entropy objective as in supervised fine-tuning, with the teacher action labels.
Optionally, we can further fine-tune the router and experts jointly with RL, following the same procedure as in \cref{sec:rl-ft}.

\section{Lightweight yet Capable Architecture: ReCAT}
\label{sec:recursive}

Our MoE design is highly flexible, allowing, for example, experts with different architectures to be combined within a single policy.
However, this flexibility comes at the cost of increased inference overhead.
Unlike typical MoE architectures used in LLMs, where only a subset of experts is activated for each token, our design usually requires all experts to be evaluated for every agent, as each expert may employ its own interaction modelling among agents.
Consequently, the inference cost grows with the number and complexity of experts, making compact yet capable expert architectures particularly desirable.

This motivates us to develop a new architecture that remains compact yet highly performant.
To this end, this section introduces \emph{\modelfullname{} (\modelname)}, illustrated in \cref{fig:recat}, which integrates prominent architectural components that have proven effective across different domains.
As indicated by its name, ReCAT represents each candidate action as a token.
A spatial encoder first encodes the local observation into each action token, after which a recurrent GNN iteratively refines these tokens by exchanging messages both within and between agents.
Finally, a linear projection converts the refined tokens into action logits.

\subsection{Spatial Encoding}
\label{sec:spatial-encoder}

The spatial encoder transforms each observation $o_i$ into a $C$-channel feature map while preserving its spatial resolution, thereby retaining the correspondence between individual FOV cells and their feature representations.
We first apply a $3\times3$ convolution, followed by group normalisation (GN)~\cite{wu2018group}, which normalises each agent's features independently of other agents and the batch size, and a SiLU activation~\cite{elfwing2018sigmoid,tan2019efficientnet}, to obtain the initial feature map $F_i^{(0)}$:
\begin{equation}
  F_i^{(0)} = \operatorname{SiLU}\!\left(\operatorname{GN}
  (\operatorname{Conv}_{3\times3}(o_i))\right).
\end{equation}
Inspired by lightweight CNN architectures~\cite{howard2017mobilenets}, our encoder next applies $B_{\mathrm{enc}}$ residual blocks, each consisting of a depthwise $3\times3$ convolution, $\operatorname{DW}_{d_b}$, with dilation $d_b$ to enlarge the receptive field without reducing the spatial resolution~\cite{yu2016multi}, followed by a pointwise $1\times1$ convolution ($\operatorname{PW}$).
More concretely,
\begin{equation}
 \begin{aligned}
  U_i^{(b)} &= \operatorname{SiLU}\!\left(\operatorname{GN}
    (\operatorname{DW}_{d_b}(F_i^{(b-1)}))\right),\\
  F_i^{(b)} &= \operatorname{SiLU}\!\left(F_i^{(b-1)}
    +\operatorname{GN}(\operatorname{PW}(U_i^{(b)}))\right).
 \end{aligned}
 \label{eq:spatial-encoder}
\end{equation}
The final feature map $F_i=F_i^{(B_{\mathrm{enc}})}$ assigns a $C$-dimensional vector $F_i(v)$ to each cell $v$ in the FOV.

\subsection{Action Tokenisation}
\label{sec:action-tokens}

For each action $a$, let $v_a \in V$ denote the location reached by applying $a$ from the current location $Q[i]$.
We tokenise each action into an initial action token $h_{i,a}^{0}\in\mathbb{R}^{d}$ by applying an MLP to the concatenation of the spatial features at the current and destination cells, the resulting change in cost-to-go $\Delta c_{i, a} \defeq \dist(v_a, g_i){-}\dist(Q[i], g_i)$, and a learned action embedding:
\begin{equation}
  h^0_{i,a} = \operatorname{MLP}\!\left(
  [F_i(Q[i]), F_i(v_a), \Delta c_{i, a}, e_a]
  \right),
  \label{eq:action-token}
\end{equation}
where $e_a\in\mathbb{R}^{d_a}$ is the action embedding.
The MLP comprises two linear layers of width $d$, with a SiLU activation between them and layer normalisation~\cite{ba2016layer} after the second layer.

\subsection{Recurrent Token Update}
\label{sec:action-graph}

We next recursively refine the action tokens while capturing interactions among agents.
Inspired by looped Transformers~\cite{dehghani2019universal,yang2024looped} and recurrent GNNs~\cite{li2016gated}, we repeatedly apply a shared update module, enabling multiple rounds of refinement without increasing the number of parameters.

As illustrated in \cref{fig:recat}, the refinement is performed through GNN message passing.
For this purpose, we expand the communication graph $G\sub{comm}$ used in MAGAT into an action-level graph $G\sub{act}=(V\sub{act},E\sub{act})$, whose nodes correspond to the five action tokens of each agent, i.e., $|V\sub{act}|=5n$.
Within each agent, all five action tokens are mutually connected, including self-edges, with zero edge features, whereas across agents, tokens corresponding to the same action are connected according to $G\sub{comm}$ with edge features $\omega_{ji}$, as in MAGAT.
Thus, $|E\sub{act}|=25n+5|E\sub{comm}|$.
GATv2~\cite{brody2021attentive} then performs message passing over $G\sub{act}$, and a gated recurrent unit (GRU)~\cite{cho2014learning} updates each token using the aggregated message and its hidden state from the previous round.

More formally, let $h^\ell\in\mathbb{R}^{5n\times d}$ denote the stacked token states at round $\ell$ and $\operatorname{LN}$ denote layer normalisation.
We apply the following update for $L$ rounds with shared parameters:
\begin{equation}
m^{\ell} = \operatorname{GATv2}\!\left(\operatorname{LN}(h^\ell),G\sub{act}\right),\
h^{\ell+1} = \operatorname{LN}\!\left(\operatorname{GRU}(m^\ell,h^\ell)\right).
 \label{eq:action-update}
\end{equation}

\subsection{Action Logits}
\label{sec:action-scores}
Finally, two distinct linear projections $f_0,f_L:\mathbb{R}^d\to\mathbb{R}$ map the initial token $h^0_{i,a}$ and the final token $h^L_{i,a}$ to scalar scores, with a learned coefficient $\lambda$ weighting the latter:
\begin{equation}
  z_{i,a} = f_0(h^0_{i,a})+\lambda f_L(h^L_{i,a}).
  \label{eq:action-logits}
\end{equation}
These logits yield the action distribution via softmax.

\section{Evaluation}
\label{sec:experiments}

We evaluate our MoE pipeline with both MAGAT and ReCAT.
All training and evaluation runs are conducted on a single NVIDIA H200 GPU using the POGEMA toolkit~\cite{skrynnik2025pogema}.

\subsection{Experimental Setup}
\label{sec:exp:training}

\subsubsection{Dataset}
Following prior GNN-based policy training studies~\cite{jain2026graph,jain2026hmagat}, the pretraining dataset contains $21{,}000$ instances with 20\% \mapname{Random} and 80\% \mapname{Maze} instances, where widths and heights are between 17 and 21 cells and contain 16, 32, or 48 agents.
We generate teacher trajectories with LaCAM3~\cite{okumura2024lacam3}.
The supervised fine-tuning datasets contain approximately $3{,}000$ instances per target distribution, with 48, 64, or 80 agents.
For router training, we combine the four SFT datasets, resulting in approximately $3{,}000 \times 4$ instances in total.
To exclude low-quality demonstrations, we retain solutions whose makespan, normalised by $\max_{i\in A}\dist(s_i,g_i)$, is at most $1.4$ for each target distribution.

For all training stages, we split the instances into training and validation sets in an $8{:}2$ ratio.
As teacher trajectories contain many wait actions from agents that have already reached their goals, which may weaken the learning signal, we follow~\cite{andreychuk2024mapfgpt} and mask 80\% of these action labels in the cross-entropy loss while retaining all other action labels.
The corresponding agents remain in the communication graph.
This masking excludes approximately 38\% of all action labels in the initial pretraining dataset.

\begin{figure*}[t]
  \centering
  \includegraphics[width=0.495\textwidth]{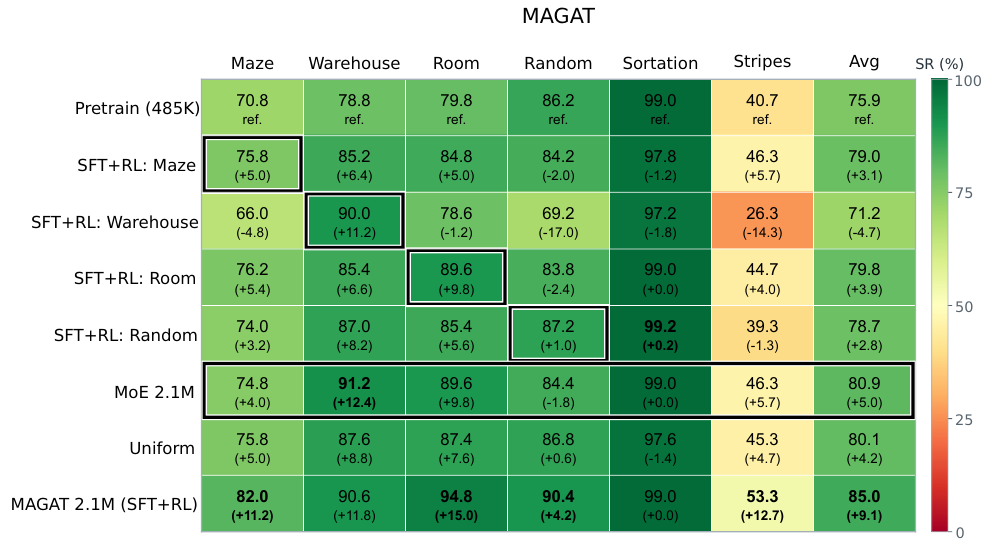}\hfill
  \includegraphics[width=0.495\textwidth]{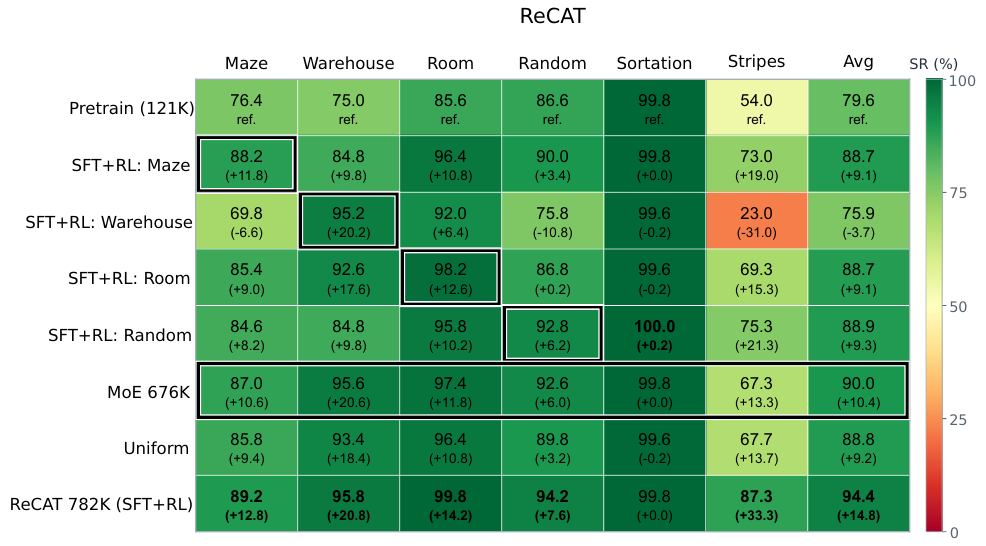}
  \par\medskip
  \input{fig/raw/evaluation}
  \captionsetup{skip=-2pt}
  \caption{
    Main results.
    \emph{Top}: success rates (\%) for MAGAT (left) and ReCAT (right); values in parentheses indicate percentage-point differences from the base pretrained policy, the Avg column denotes the unweighted mean across the six map distributions, and the Uniform row combines expert logits with equal weights.
    \emph{Bottom}: performance across seven map distributions and varying numbers of agents, comparing base pretrained policies, MoE policies, larger monolithic policies with SFT+RL, HMAGAT, and MAPF-GPT-DDG.
    Shaded bands indicate 95\% bootstrap confidence intervals for SoC.
  }
  \label{fig:evaluation-pibt}
\end{figure*}

\subsubsection{Training Procedure}
We use AdamW~\cite{loshchilov2019adamw} for pretraining, SFT, and router training, with weight decay of $0.05$ and cosine annealing of the learning rate to $10^{-6}$.
For pretraining and SFT, we adjust the batch size according to the fraction of retained labels to maintain an effective batch size of approximately 48.

\paragraph{Pretraining}
Pretraining runs for 200 epochs with an initial learning rate of $10^{-3}$.
Following~\cite{jain2026hmagat}, we perform on-demand data aggregation every four epochs by running the current policy with naive collision shielding on 500 instances.
For each failed rollout, we invoke LaCAM3 from its final configuration and add the resulting teacher trajectory to the training dataset.
Furthermore, inspired by~\cite{andreychuk2025finetuning}, once the rollout success rate reaches 80\%, we additionally revisit successful rollouts whose makespan is at least $1.2$ times that of the teacher trajectory.
For up to 30 such instances per aggregation round, we sample configurations at 16-timestep intervals and invoke LaCAM3 from each sampled configuration.
We add the resulting teacher trajectory if its makespan is at most $1/1.2$ of the remaining policy rollout.
Pretraining takes approximately 14 hours and 21 hours for the base MAGAT and ReCAT models, respectively, including validation and on-demand data aggregation.

\paragraph{SFT}
It runs for 52 epochs with an initial learning rate of $10^{-5}$, using only the pre-collected trajectories for the target distribution without on-demand data aggregation, taking a few hours per expert.

\paragraph{RL Fine-Tuning}
Each supervised expert undergoes 100 epochs of RL on its target map distribution.
In each epoch, we sample 25 instances with 48, 64, or 80 agents and collect $B{=}4$ rollouts per instance, each capped at $T\sub{max}{=}128$ timesteps.
We then perform three optimisation passes over minibatches of the collected rollouts using Adam~\cite{kingma2015adam} with a learning rate of $10^{-6}$, a clipping threshold $\epsilon=0.2$, reference-policy KL coefficient $\beta=0.01$, entropy coefficient $\alpha=0.001$, and the gradient norm clipped at $0.5$.

\paragraph{Router Training}
With the expert weights frozen, we train the router for 30 epochs with an initial learning rate of $10^{-3}$ and a fixed routing temperature of $\tau=1$.
Router training takes approximately 2.5 hours for MAGAT-MoE and 5.5 hours for \modelname{}-MoE.

\subsubsection{Model Configurations}
We use the same observation representation and communication graph for MAGAT and ReCAT, with an observation radius of $R\sub{obs}=5$ and a communication radius of $R\sub{comm}=5$.
Our base MAGAT has 485K parameters and consists of three stacked GNN layers.
Our base ReCAT has 121K parameters, with $B\sub{enc}=4$ residual spatial encoder blocks with $C=64$ channels, action tokens of dimension $d=96$, and $L=4$ communication rounds.
The MLP router has 190K parameters.

\subsubsection{Baselines}
We compare \textbf{MAGAT-MoE} (2.1M parameters) and \textbf{ReCAT-MoE} (676K), with routers trained by supervised learning while the experts remain frozen, against their respective pretrained policies and distribution-specific experts obtained through SFT and RL fine-tuning, to assess how well the mixture preserves specialist performance.
We also evaluate \textbf{Uniform}, which combines the same four frozen SFT+RL experts by averaging their action logits before applying softmax to rate the learned routing efficacy.

For comparison with existing methods, we consider two leading neural MAPF policies.
\textbf{HMAGAT} (1M)~\cite{jain2026hmagat} is an IL policy that models agent interactions using hypergraph neural layers and employs RL to adapt each agent's action-sampling temperature.
\textbf{MAPF-GPT-DDG} (2M)~\cite{andreychuk2025finetuning} is a compact variant of the Transformer-based MAPF-GPT policy~\cite{andreychuk2024mapfgpt} that achieves comparable performance with a smaller model through targeted expert data augmentation.
For these, we use the authors' official code repositories.

We also prepare parameter-matched monolithic versions of \textbf{MAGAT} (2.1M) and \textbf{ReCAT} (782K) by increasing their latent dimensions accordingly.
We first pretrain these models on the same instances as the smaller pretrained base models, taking approximately 15 hours for MAGAT and 30 hours for ReCAT.
For a fair comparison, as the MoE policies are exposed to instances with larger numbers of agents during training, these monolithic models subsequently undergo SFT on the concatenation of the four SFT datasets, using the same settings as expert training.
RL fine-tuning follows the expert training procedure, with the four training distributions sampled with equal probability.
SFT takes approximately 5 hours for MAGAT and 10 hours for ReCAT, while RL fine-tuning takes approximately 8 hours per policy.

All main evaluations use CS-PIBT, given its substantial performance improvement over naive collision shielding.

\subsubsection{Metrics and Instances}
We report the success rate (all agents reaching their goals) and mean relative SoC with respect to LaCAM3 solutions obtained with a \SI{60}{\second} time budget.
For unsuccessful instances, agents that have not reached their goals are assigned the timestep limit as their individual costs when computing SoC.
For in-distribution evaluation (\mapname{Maze}, \mapname{Warehouse}, \mapname{Room}, and \mapname{Random}), there is no overlap between training and test instances.
For out-of-distribution evaluation, we use a fixed map layout for each of \mapname{Sortation}, derived from~\cite{chen2024traffic}, and \mapname{Stripes}, from~\cite{jain2026graph}.
We also generate obstacle-free \mapname{Empty} maps with widths and heights between 17 and 21 cells; see \cref{fig:evaluation-pibt}.
For each agent count, we use the same 100 fixed instances, each with a limit of 128 timesteps.

\subsection{Main Results}
\Cref{fig:evaluation-pibt} summarises success rates and mean relative SoC, while \cref{tab:pibt-comparison} reports aggregated scores over six map distributions with and without CS-PIBT.
Several key observations highlight the benefits of MoE across both underlying architectures.

\paragraph{Map-distribution Experts}
The success-rate heatmaps show that our fine-tuning strategy substantially improves policy performance when the training and test distributions match, but these gains do not transfer uniformly to other distributions.
For example, for both architectures, the \mapname{Warehouse} expert achieves a higher success rate than the pretrained policy on \mapname{Warehouse}, but lower success rates on \mapname{Maze} and \mapname{Random}.
This indicates that \emph{the resulting fine-tuned policies indeed become specialists for specific map distributions.}

\paragraph{Efficacy of MoE}
Unlike individual experts, whose success rates can drop substantially on certain distributions, \emph{MoE policies retain the experts' strong in-distribution performance while generally improving upon the pretrained models across distributions.}
MoE generally outperforms Uniform too, indicating that the learned router effectively exploits complementary strengths across experts rather than merely averaging their outputs.
Furthermore, despite using fewer parameters, ReCAT-MoE outperforms the leading HMAGAT and MAPF-GPT-DDG baselines across all distributions when considering success rate and solution cost jointly.
MAGAT-MoE also achieves comparable aggregate performance to both baselines in terms of success rate and solution cost, further demonstrating the effectiveness of our training pipeline.
\Cref{tab:pibt-comparison} also shows that the benefits of the learned mixtures extend to the setting without CS-PIBT.
Nevertheless, as seen for ReCAT on \mapname{Stripes}, the MoE policy does not always match the best-performing expert for each map distribution, leaving room for improvement in expert utilisation.

\paragraph{Expert Usage Analysis}
\Cref{tab:expert-usage} reports the mean mixing weight assigned to each expert during evaluation.
For both architectures, the expert trained on the corresponding distribution receives the largest weight on each in-distribution map, even though the router is not given a map label.
On all three out-of-distribution maps, the \mapname{Maze} expert receives the largest weight.
Still, the router assigns weight to the other experts too, combining their specialised outputs rather than relying exclusively on a single expert.

\paragraph{Does MoE offer benefits over monolithic policies?}
While both MoE policies achieve strong performance, their parameter-matched monolithic counterparts reveal a more nuanced picture;
the monolithic policies, trained using the same pretraining--SFT--RL recipe, outperform the MoE models.
In particular, the larger ReCAT, which incorporates effective components adopted across the literature, achieves the best performance among all evaluated methods by a substantial margin.
This result suggests that \emph{designing an effective underlying architecture under the same parameter budget can yield more powerful policies than map-distribution-based MoE}.

{
\begin{table}[t]
\centering
\caption{
Aggregated scores with and without CS-PIBT over the same 2,800 instances across six map distributions.
SR denotes success rate; relative SoC and runtime are means over all instances, including failures.
}
\label{tab:pibt-comparison}
\small
\setlength{\tabcolsep}{4pt}
\renewcommand{\arraystretch}{1.15}
\begin{tabular*}{\linewidth}{@{\extracolsep{\fill}}lrrrrrr@{}}
\toprule
& \multicolumn{3}{c}{CS-PIBT} & \multicolumn{3}{c}{w/o CS-PIBT} \\
\cmidrule(lr){2-4}\cmidrule(lr){5-7}
Model & SR (\%) & Rel. SoC & Time (\si{\second}) & SR (\%) & Rel. SoC & Time (\si{\second}) \\
\midrule
\solver{MAGAT} (485K) & 78.4 & 1.752 & 0.879 & 62.4 & 2.161 & 0.501 \\
\solver{MAGAT} Uniform & 82.6 & 1.578 & 1.073 & 67.6 & 2.017 & 0.778 \\
\solver{MAGAT} MoE & 83.4 & 1.555 & 1.173 & 67.8 & 1.984 & 0.821 \\
\solver{MAGAT} (2.1M, SFT+RL) & 87.3 & 1.470 & 0.733 & 73.6 & 1.829 & \textbf{0.412} \\
\modelname (121K) & 81.4 & 1.664 & 0.921 & 69.5 & 1.981 & 0.535 \\
\modelname Uniform & 90.3 & 1.414 & 1.252 & 76.9 & 1.729 & 1.042 \\
\modelname MoE & 91.6 & 1.402 & 1.241 & 78.0 & 1.730 & 1.061 \\
\modelname (782K, SFT+RL) & \textbf{94.9} & \textbf{1.313} & \textbf{0.723} & \textbf{85.6} & \textbf{1.520} & 0.437 \\
HMAGAT & 80.1 & 1.483 & 2.069 & 37.4 & 2.191 & 2.432 \\
MAPF-GPT-DDG & 80.2 & 1.749 & 1.310 & 55.0 & 2.455 & 1.072 \\
\bottomrule
\end{tabular*}
\end{table}
}

{
\setlength{\tabcolsep}{4pt}
\renewcommand{\arraystretch}{1.2}
\begin{table}[!t]
\centering
\caption{
Mean expert mixture weights (\%).
Columns correspond to evaluation distributions.
}
\label{tab:expert-usage}
\small
\begin{tabular*}{\linewidth}{@{\extracolsep{\fill}}llrrrrrrrr@{}}
\toprule
&Expert &\mapname{Maze} & \mapname{WH.} & \mapname{Room} & \mapname{Random} & \mapname{Sort.} & \mapname{Stripes} & \mapname{Empty} & Avg. \\
\midrule
\multirow{4}{*}{\rotatebox{90}{\footnotesize MAGAT-MoE}}
&Maze & \textbf{97.4} & 11.0 & 12.4 & 10.1 & \textbf{63.3} & \textbf{94.5} & \textbf{65.5} & \textbf{50.6} \\
&WH. & 0.3 & \textbf{78.9} & 0.5 & 1.1 & 0.7 & 0.0 & 12.1 & 13.4 \\
&Room & 1.0 & 1.7 & \textbf{83.3} & 2.5 & 18.4 & 5.4 & 0.6 & 16.1 \\
&Random & 1.2 & 8.4 & 3.9 & \textbf{86.3} & 17.5 & 0.1 & 21.8 & 19.9 \\
\midrule
\multirow{4}{*}{\rotatebox{90}{\footnotesize \modelname{}-MoE}}
&Maze & \textbf{70.4} & 22.0 & 28.2 & 25.0 & \textbf{49.7} & \textbf{69.5} & \textbf{42.4} & \textbf{43.9} \\
&WH. & 8.7 & \textbf{72.6} & 8.1 & 6.5 & 9.0 & 9.1 & 18.9 & 19.0 \\
&Room & 11.7 & 1.4 & \textbf{57.0} & 5.0 & 26.3 & 18.6 & 0.5 & 17.2 \\
&Random & 9.2 & 4.0 & 6.8 & \textbf{63.5} & 15.0 & 2.8 & 38.2 & 19.9 \\
\bottomrule
\end{tabular*}%
\end{table}
}

{
\setlength{\tabcolsep}{4pt}
\renewcommand{\arraystretch}{1.15}
\begin{table}[!t]
\centering
\caption{
Expert performance on \mapname{Warehouse} and \mapname{Maze} across different training strategies.
PL denotes the Plackett--Luce likelihood of the candidate sequence examined by PIBT; w/o PL uses only the selected action's probability for RL updates.
}
\label{tab:warehouse-posttraining}
\small
\begin{tabular*}{0.9\linewidth}{@{\extracolsep{\fill}}llrrrr@{}}
\toprule
& & \multicolumn{2}{c}{\mapname{Warehouse}} & \multicolumn{2}{c}{\mapname{Maze}} \\
\cmidrule(lr){3-4}\cmidrule(lr){5-6}
& Training & SR (\%) $\uparrow$ & Rel. SoC $\downarrow$ & SR (\%) $\uparrow$ & Rel. SoC $\downarrow$ \\
\midrule
\multirow{6}{*}{\rotatebox{90}{\footnotesize MAGAT (485K)}}
& Pretrain & 78.8 & 1.900 & 70.8 & 1.826 \\
& SFT & 85.6 & 1.646 & 74.4 & 1.711 \\
& RL (w/o PL) & 83.8 & 1.928 & 75.0 & 1.854 \\
& RL (PL) & 87.2 & 1.792 & 77.4 & 1.707 \\
& SFT+RL (w/o PL) & \textbf{90.0} & 1.649 & \textbf{80.4} & 1.708 \\
& SFT+RL (PL) & \textbf{90.0} & \textbf{1.602} & 75.8 & \textbf{1.630} \\
\midrule
\multirow{6}{*}{\rotatebox{90}{\footnotesize \modelname (121K)}}
& Pretrain & 75.0 & 1.838 & 76.4 & 1.738 \\
& SFT & 90.4 & 1.522 & 82.4 & 1.568 \\
& RL (w/o PL) & 94.2 & 1.649 & 81.8 & 1.642 \\
& RL (PL) & 89.2 & 1.645 & 85.2 & 1.549 \\
& SFT+RL (w/o PL) & \textbf{95.4} & 1.438 & 85.4 & 1.607 \\
& SFT+RL (PL) & 95.2 & \textbf{1.400} & \textbf{88.2} & \textbf{1.472} \\
\bottomrule
\end{tabular*}
\end{table}
}

{
\begin{table}[!t]
\centering
\caption{Scalability assessment.
\textit{Times} report network inference.
The versions with CS-PIBT is shown with parentheses (in Python unlike \cite{jiang2025deploying,jain2026graph} using C++).
\textit{Goal} denotes the percentage of agents at their goals.
\emph{SoC} is normalised by $\sum_{i \in A} \dist(s_i, g_i)$.
}
\label{tab:scalability}
  \scriptsize
  \setlength{\tabcolsep}{1pt}
  \renewcommand{\arraystretch}{1.15}
\resizebox{\linewidth}{!}{%
\begin{tabular}{@{}r*{15}{r}@{}}
\toprule
& \multicolumn{3}{c}{MAGAT-MoE (2.1M)} & \multicolumn{3}{c}{MAGAT (2.1M)} & \multicolumn{3}{c}{\modelname{}-MoE (676K)} & \multicolumn{3}{c}{\modelname{} (782K)} & \multicolumn{3}{c}{MAPF-GPT-DDG (2M)} \\
\cmidrule(lr){2-4}\cmidrule(lr){5-7}\cmidrule(lr){8-10}\cmidrule(lr){11-13}\cmidrule(l){14-16}
\# Agents & ms/step & Goal (\%) & SoC & ms/step & Goal (\%) & SoC & ms/step & Goal (\%) & SoC & ms/step & Goal (\%) & SoC & ms/step & Goal (\%) & SoC \\
\midrule
1,000 & 5.9 (\textbf{79.9}) & 100 & 1.017 & \textbf{1.8} (81.8) & 100 & 1.015 & 11.4 (83.5) & 100 & 1.019 & 6.3 (85.7) & 100 & \textbf{1.013} & 40.0 (115.3) & 100 & 1.032 \\
10,000 & 13.1 (727.7) & 100 & 1.095 & \textbf{5.0} (\textbf{707.1}) & 100 & 1.090 & 97.7 (809.9) & 100 & 1.080 & 60.3 (771.1) & 100 & \textbf{1.079} & 424.2 (1145.2) & 100 & 1.100 \\
100,000 & 620.6 (7402.2) & 97.122 & 2.318 & \textbf{285.6} (\textbf{7068.5}) & 98.335 & 2.233 & 2189.8 (9022.2) & 99.466 & 2.104 & 1227.5 (8191.5) & \textbf{99.711} & \textbf{2.020} & 4476.2 (11435.6) & 96.555 & 2.415 \\
\bottomrule
\end{tabular}%
}
\end{table}
}

{
\newcommand{\frozenicon}{%
  \tikz[x=0.9em,y=0.9em,baseline=-0.6ex,line width=0.45pt,line cap=round]{%
    \foreach \angle in {0,60,...,300}{%
      \draw (0,0) -- (\angle:0.48);
      \draw[rotate=\angle] (0.31,0.08) -- (0.41,0) -- (0.31,-0.08);
    }
  }%
}
\newcommand{\updatedicon}{%
  \tikz[x=0.9em,y=0.9em,baseline=-0.6ex]{%
    \fill (0,-0.48)
      .. controls (-0.39,-0.48) and (-0.52,-0.12) .. (-0.32,0.17)
      .. controls (-0.34,-0.02) and (-0.15,-0.06) .. (-0.16,0.12)
      .. controls (-0.17,0.30) and (0.07,0.37) .. (0.02,0.54)
      .. controls (0.36,0.33) and (0.15,0.14) .. (0.31,0.04)
      .. controls (0.39,0.08) and (0.41,0.15) .. (0.39,0.22)
      .. controls (0.59,-0.15) and (0.39,-0.48) .. cycle;
    \fill[white] (0,-0.36)
      .. controls (-0.18,-0.34) and (-0.20,-0.14) .. (-0.03,0.04)
      .. controls (-0.04,-0.12) and (0.19,-0.16) .. (0.14,-0.28)
      .. controls (0.11,-0.35) and (0.06,-0.36) .. cycle;
  }%
}
\begin{table}[t]
\centering
\caption{
Performance across different MoE training strategies w.r.t. success-rate gains over the pretrained ReCAT (121K), in percentage points.
Icons: frozen \protect\frozenicon~or updated~\protect\updatedicon~parameters.
}
\label{tab:moe-training}
\small
\setlength{\tabcolsep}{3pt}
\renewcommand{\arraystretch}{1.15}
\begin{tabular*}{\linewidth}{@{\extracolsep{\fill}}ccccrrrrrrr@{}}
\toprule
\multicolumn{2}{c}{Supervised training} & \multicolumn{2}{c}{RL} & \multirow{2}{*}{\mapname{Maze}} & \multirow{2}{*}{\mapname{WH.}} & \multirow{2}{*}{\mapname{Room}} & \multirow{2}{*}{\mapname{Random}} & \multirow{2}{*}{\mapname{Sort.}} & \multirow{2}{*}{\mapname{Stripes}} & \multirow{2}{*}{Avg.} \\
\cmidrule(lr){1-2}\cmidrule(lr){3-4}
Experts & Router & Experts & Router & & & & & & & \\
\midrule
\frozenicon & \updatedicon & \multicolumn{2}{c}{N/A} & +10.6 & +20.6 & +11.8 & +6.0 & +0.0 & +13.3 & +10.4 \\
\updatedicon & \updatedicon & \multicolumn{2}{c}{N/A} & +6.0 & +15.8 & +9.8 & +3.8 & -0.4 & -18.3 & +2.8 \\
\frozenicon & \updatedicon & \frozenicon & \updatedicon & +9.0 & +21.2 & +12.6 & +6.6 & -0.2 & +16.0 & +10.9 \\
\frozenicon & \updatedicon & \updatedicon & \updatedicon & \textbf{+13.0} & \textbf{+21.6} & \textbf{+12.8} & \textbf{+8.4} & \textbf{+0.2} & \textbf{+24.3} & \textbf{+13.4} \\
\updatedicon & \updatedicon & \frozenicon & \updatedicon & +3.6 & +17.0 & +9.2 & +5.4 & -0.4 & -18.3 & +2.7 \\
\updatedicon & \updatedicon & \updatedicon & \updatedicon & +8.0 & +19.0 & +11.0 & +6.4 & \textbf{+0.2} & -5.3 & +6.5 \\
\midrule
\multicolumn{4}{l}{\modelname (782K, SFT+RL)} & +12.8 & +20.8 & +14.2 & +7.6 & +0.0 & +33.3 & +14.8 \\
\bottomrule
\end{tabular*}%
\end{table}
}

\subsection{Large-Scale Experiment}
We next evaluate these models on substantially larger instances with up to $100{,}000$ agents on larger $512{\times}512$ \mapname{Random} maps with 20\% obstacles, with start--goal distances at most 64.
\Cref{tab:scalability} shows that both \{MAGAT, ReCAT\}-MoE achieve substantially faster inference than the baseline MAPF-GPT-DDG, while HMAGAT cannot operate at this scale due to the computational overhead of hypergraph construction.
Moreover, ReCAT-MoE achieves superior solution performance while retaining this computational scalability.
Still, the monolithic policies achieve faster inference and better solution quality.

\subsection{Expert Training Ablation}
The performance of our MoE policies naturally depends on the quality of the underlying experts, making an effective expert training recipe as important as the MoE training itself.
\Cref{tab:warehouse-posttraining} investigates this aspect by focusing on \mapname{Warehouse} and \mapname{Maze}.
For both architectures, either SFT or RL improves policy performance in terms of both success rate and solution quality, while combining them yields the best overall results, supporting our expert training strategy.
The table also ablates the Plackett--Luce likelihood used during GRPO fine-tuning, showing that it consistently reduces SoC compared with using only the selected action's probability, although its effect on success rate varies across architectures and map distributions.

\subsection{MoE Training Ablation}
We next examine the MoE training pipeline, using ReCAT.
Thus far, our MoE models use frozen experts and train only the router through supervised learning.
However, the experts can also be jointly optimised during router training, and the resulting MoE can be further fine-tuned with RL.
\Cref{tab:moe-training} compares these alternative training strategies after expert preparation.

The result shows that keeping the experts frozen during supervised training generally yields larger gains than updating them jointly with the router.
Among the evaluated variants, supervised router training with frozen experts, followed by RL for both the router and experts, yields the largest gains;
this matches the monolithic model on most distributions, although it still falls behind on average, mainly due to the out-of-distribution scenario \mapname{Stripes}.

\section{Discussion}
Across two distinct policy architectures, this study demonstrates that MoE over map-distribution-specialised experts provides a capable approach to neural MAPF.
In particular, despite its compact construction, ReCAT-MoE achieves strong performance in both in-distribution and out-of-distribution settings.
However, we also find that, under the same parameter budget, monolithic policies still provide stronger performance, suggesting that realising the benefits of MoE---a paradigm that has proven highly successful in recent LLMs---remains challenging in the MAPF context.
We believe that alternative MoE architectures warrant further investigation.
In the following, we position our study within the broader literature and discuss its limitations and future directions.

\paragraph{MoE}
MoE~\cite{jacobs1991adaptive} is a long-established concept in machine learning that combines multiple experts through a routing function.
Broadly, MoE training can follow two approaches: learning a router over frozen experts or jointly learning the router and experts.
While recent LLMs predominantly follow the latter approach~\cite{shazeer2017outrageously,dai2024deepseekmoe}, our approach largely falls into the former, leveraging the fact that map-wise fine-tuning provides strong specialised experts for MAPF~\cite{jain2026graph}.
Meanwhile, the superior performance obtained by jointly fine-tuning the experts and router with RL suggests that expert specialisation may also emerge implicitly through joint learning.
This points to a key limitation of our current approach, i.e., the need to explicitly construct experts for predefined map distributions, while suggesting that learning expert specialisation implicitly may open a different avenue for MoE in neural MAPF.

\paragraph{Multiple Algorithms and Policies in MAPF}
We clarify that this is not the first study to utilise multiple algorithms or policies in MAPF.
For example, algorithm selection~\cite{kaduri2020algorithm,ren2021mapfast} aims to choose a suitable algorithm from a portfolio of MAPF algorithms for a given target instance; by applying the selected solver, this meta-scheme aims to find solutions efficiently.
Tang et al.~\cite{tang2024ensembling} adopt best-trajectory sampling over multiple policies configured with different hyperparameters.
Other approaches switch between policies according to predefined rules~\cite{okumura2022priority,gao2023rde,jain2026graph}, or combine collision-avoidance and goal-reaching policies by additively combining their outputs, similarly to potential-field methods~\cite{skrynnik2021hybrid,pertzovskiy2026enhancing}.
In contrast, our contribution lies in an explicit MoE formulation that uses a learned router to adaptively fuse expert outputs according to each agent's local situation, together with an effective recipe for creating strong specialised experts.
This adaptive fusion during rollout is critical, as demonstrated by the results in \cref{fig:evaluation-pibt}.

\paragraph{ReCAT}
As described earlier, our newly designed compact neural MAPF policy, ReCAT, integrates several techniques that have proven effective in other domains.
Meanwhile, concurrent preprints have also explored recursive refinement for interaction modelling in neural MAPF policies~\cite{vyaltsev2026learning,vyaltsev2026decentralized}, similarly showing promising results.

\paragraph{Automatic Map Generation}
Our MoE analysis implicitly suggests that certain map distributions are more useful than others for training capable policies, motivating another interesting direction: automatically generating map distributions for training.
Such automatic curriculum generation has proven effective in RL~\cite{phan2025generative}, while map generation itself is an active area of MAPF research~\cite{qian2024qd}.
Integrating these techniques into neural MAPF policy training represents an exciting direction for future work.

\section*{Acknowledgments}
This research was conducted by AIST and under FRONTia, a Japanese national program led by the Ministry of Economy, Trade and Industry of Japan (METI) and the New Energy and Industrial Technology Development Organization (NEDO), aimed at developing a domestic multimodal foundation model for AI robots and physical AI.
\bibliographystyle{unsrtnat}
\bibliography{sty/ref-macro,ref}
\clearpage
\appendix
\section*{Appendix}

\section{Map Distributions}
\label{sec:map-distributions}

\begin{figure}[!ht]
  \centering
  \setlength{\tabcolsep}{3pt}
  \begin{tabular}{@{}cccc@{}}
    \mapname{Maze} & \mapname{Warehouse} & \mapname{Room} & \mapname{Random} \\[2pt]
    \includegraphics[height=0.12\columnwidth]{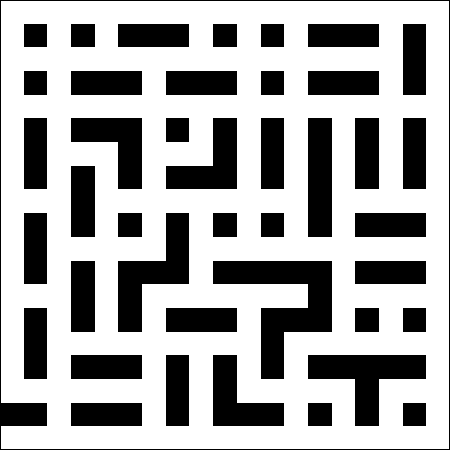} &
    \includegraphics[height=0.12\columnwidth]{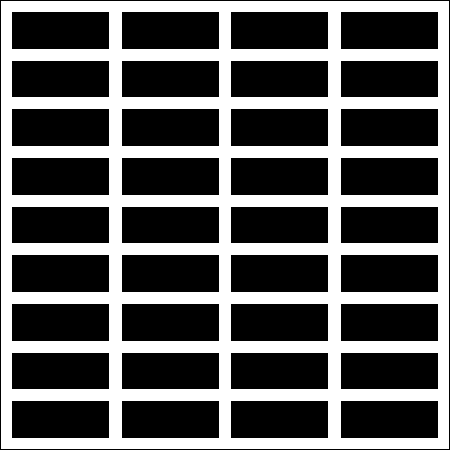} &
    \includegraphics[height=0.12\columnwidth]{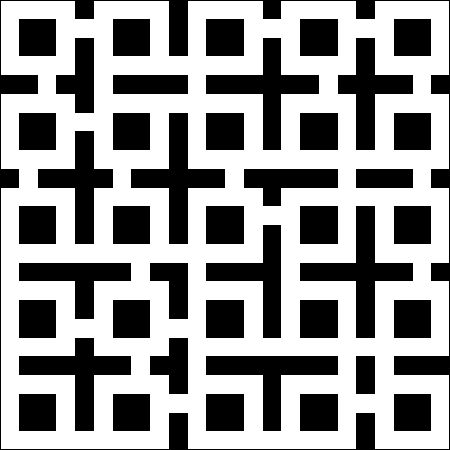} &
    \includegraphics[height=0.12\columnwidth]{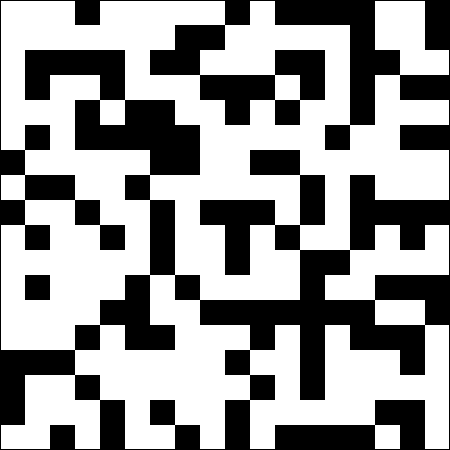} \\[2pt]
    \includegraphics[height=0.12\columnwidth]{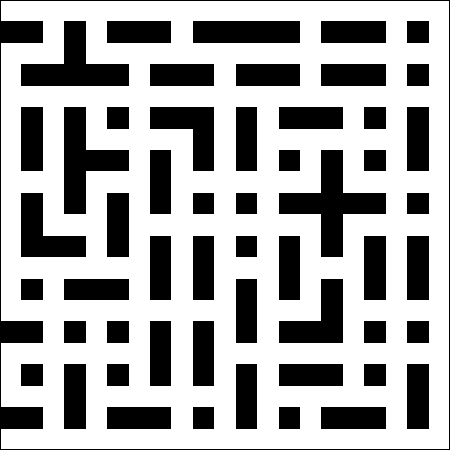} &
    \includegraphics[height=0.12\columnwidth]{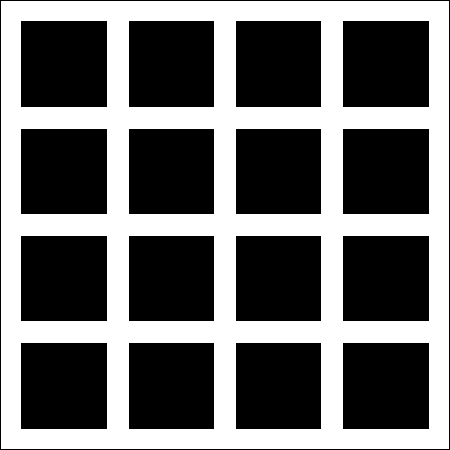} &
    \includegraphics[height=0.12\columnwidth]{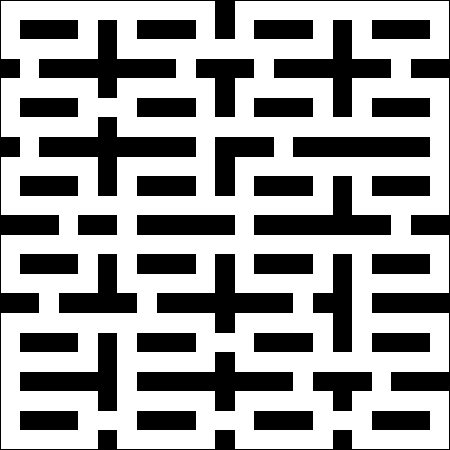} &
    \includegraphics[height=0.12\columnwidth]{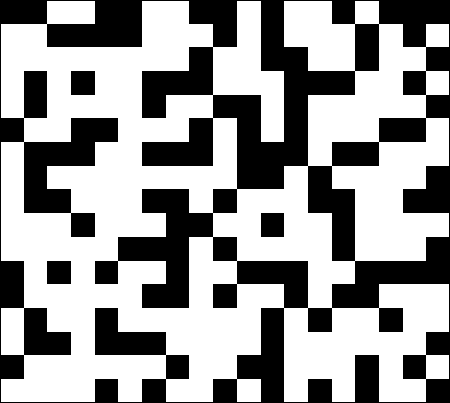} \\[2pt]
    \includegraphics[height=0.12\columnwidth]{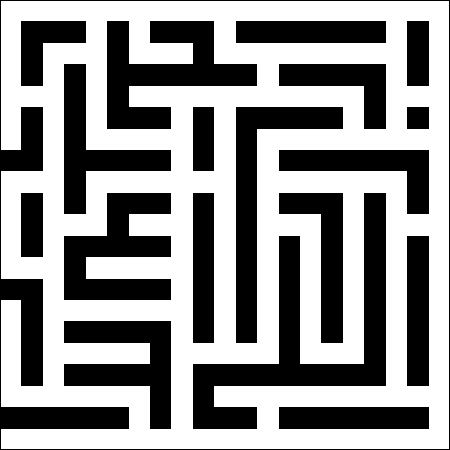} &
    \includegraphics[height=0.12\columnwidth]{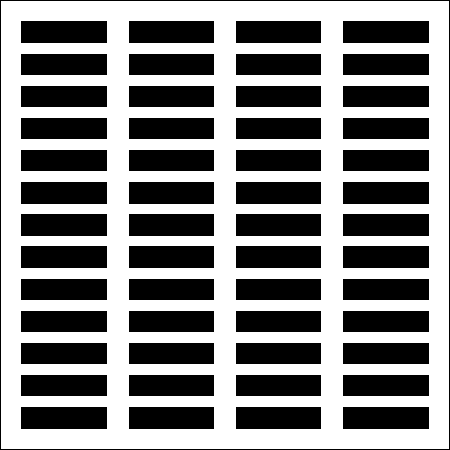} &
    \includegraphics[height=0.12\columnwidth]{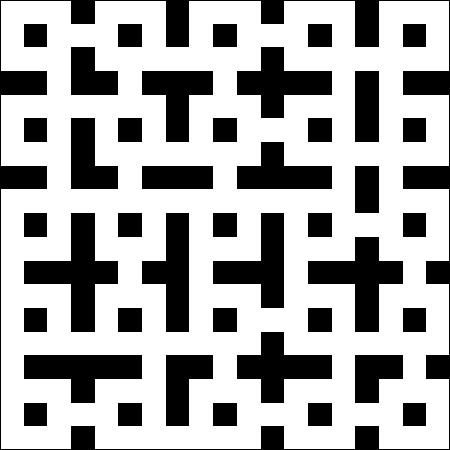} &
    \includegraphics[height=0.12\columnwidth]{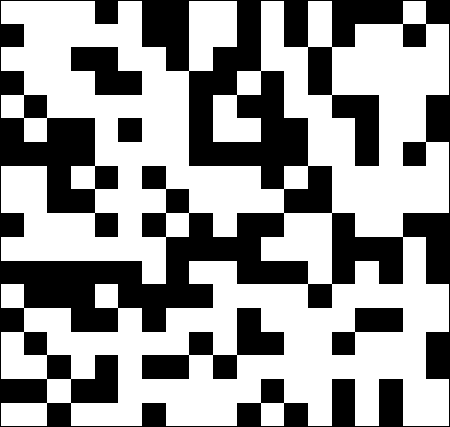} \\[2pt]
    \includegraphics[height=0.12\columnwidth]{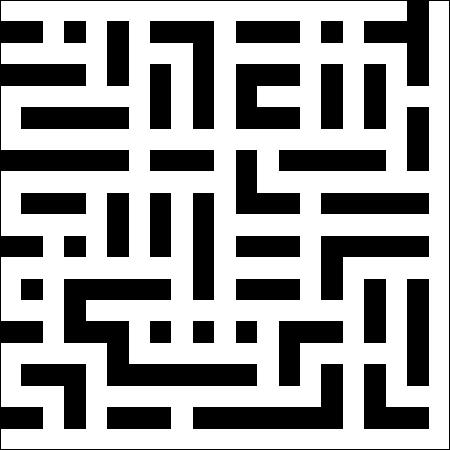} &
    \includegraphics[height=0.12\columnwidth]{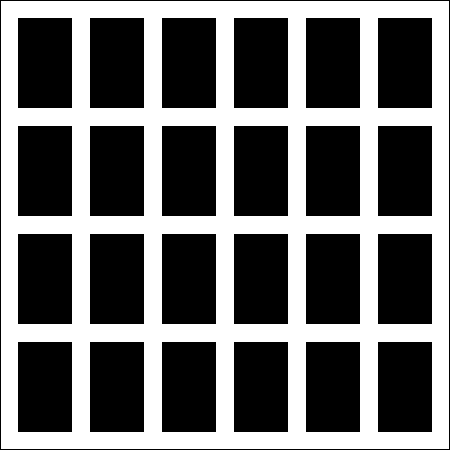} &
    \includegraphics[height=0.12\columnwidth]{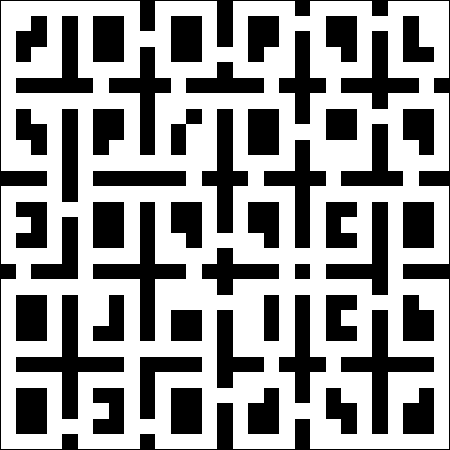} &
    \includegraphics[height=0.12\columnwidth]{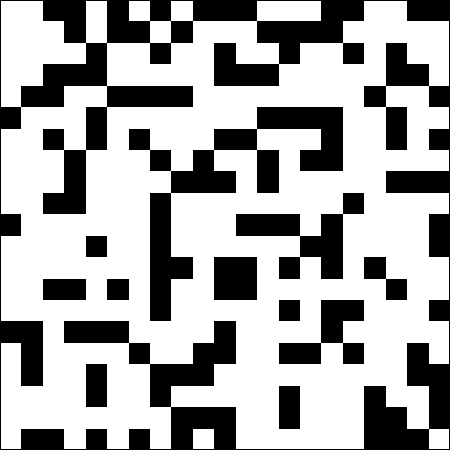}
  \end{tabular}
  \caption{Examples from the four in-distribution map generators.
  Each column shows four different layouts; black cells denote obstacles and white cells are traversable.}
  \label{fig:map-examples}
\end{figure}

\Cref{fig:map-examples} shows four layouts from each in-distribution map generator.
Start--goal pairs are sampled within connected components.
Maps that cannot accommodate the requested number of agents are resampled.
Generator parameters are sampled uniformly within the ranges below.

\paragraph{Maze}
We generate maze-like layouts by growing walls from randomly selected cells in two-cell steps, filling the intervening cell.
Even map dimensions are rounded up to the next odd integer.
Each wall seed receives 1--8 extension attempts, with a straight-continuation weight of $0.75$--$0.85$ to favour long corridors.
The density parameter, sampled between $0.2$ and $1.0$, controls the number of wall-growth attempts.

\paragraph{Warehouse}
We arrange rectangular shelves in a regular grid, with shelf widths of 2--8 cells and heights of 2--6 cells.
Aisles span 1--2 cells between shelf columns and one cell between rows, with side padding of 1--3 cells.
Map dimensions are determined from an initial arrangement of 1--4 shelf rows and columns.

\paragraph{Room}
We generate a grid with 3--6 rows and columns of rooms, whose shared width and height are independently sampled between three and five cells.
Adjacent rooms are separated by one-cell-thick walls, each with a randomly placed doorway.
A one-cell-wide perimeter remains free in each room, while interior cells are independently blocked with a probability sampled between $0.8$ and $1.0$ per map.

\paragraph{Random}
Obstacles are placed randomly, with an obstacle density sampled between 20\% and 40\%.

\clearpage

\section{Model Details}
\label{sec:model-details}

\Cref{tab:model-parameters} lists the exact parameter counts.

\begin{table}[ht]
\centering
\captionsetup{width=0.8\linewidth}
\caption{Parameter counts of the evaluated MAGAT and \modelname configurations, including all expert parameters in each mixture.}
\label{tab:model-parameters}
\small
\renewcommand{\arraystretch}{1.15}
\begin{tabular*}{0.6\linewidth}{@{\extracolsep{\fill}}lr@{}}
\toprule
Model & Parameters \\
\midrule
MAGAT (485K) & 484,933 \\
MAGAT (2.1M) & 2,154,181 \\
\modelname (121K) & 121,299 \\
\modelname (782K) & 781,843 \\
\midrule
MLP router (190K) & 190,980 \\
MAGAT-MoE (2.1M) & 2,130,712 \\
\modelname{}-MoE (676K) & 676,176 \\
MAGAT Uniform & 1,939,732 \\
\modelname Uniform & 485,196 \\
\bottomrule
\end{tabular*}
\end{table}

\subsection{MAGAT}
The encoder uses five $3\times3$ convolutions with channel widths $(32,32,64,64,128)$ and ReLU, with $2\times2$ max pooling after the first, third, and fifth layers.
A linear projection produces 256-dimensional agent embeddings.
For MAGAT (485K), the first two GATv2 layers concatenate four 16-dimensional heads, and the final layer averages four 256-dimensional heads.
The encoder embedding is added to the GAT output before an MLP decoder with two 256-dimensional hidden layers and ReLU.
MAGAT (2.1M) increases the head dimension in the first two GATv2 layers from 16 to 128, leaving the remaining architecture unchanged.

\subsection{\modelname}
\modelname (121K) and \modelname (782K) use depthwise-convolution dilation factors $(1,2,1,2)$, eight groups for group normalisation, and action embeddings of dimension $d_a=16$.
GATv2 concatenates four attention heads of dimension $d/4$, with attention dropout set to zero.
The learned coefficient $\lambda$ is initialised to $0.05$.
The larger \modelname (782K) increases the spatial channels and action-token dimension to $C=d=224$, retaining the remaining settings of the base model.

\clearpage
\section{Evaluation without CS-PIBT}
\label{sec:evaluation-nopibt}

\begin{figure}[!ht]
  \centering
  \input{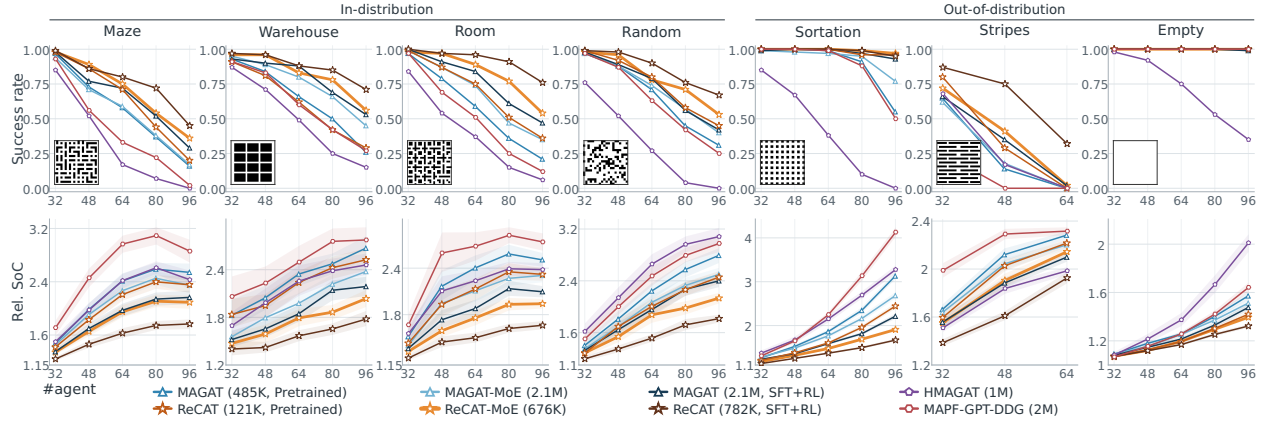}
  \captionsetup{skip=-4pt}
  \caption{
    Success rate and mean relative SoC without CS-PIBT across seven map distributions.
    Each point averages 100 instances; SoC is reported relative to LaCAM3 with a \SI{60}{\second} time limit.
    Shaded bands indicate 95\% bootstrap confidence intervals for SoC.
  }
  \label{fig:evaluation-nopibt}
\end{figure}

\Cref{fig:evaluation-nopibt} reports results with naive collision shielding on the same test instances as \cref{fig:evaluation-pibt}.

\section{Expert Exclusion}
\label{sec:expert-exclusion}

We examine whether each expert contributes to MoE performance on its target distribution.
We remove one distribution-specific expert and train a new router on the SFT data from all four distributions, keeping the remaining experts frozen.
\Cref{tab:expert-exclusion} compares these variants with the full four-expert MoE.
Removing an expert reduces success rates on its target distribution for all four \modelname{} experts and for the Warehouse and Room experts in MAGAT, highlighting the contribution of distribution-specific experts.

{
\begin{table}[!ht]
\centering
\caption{
Success rates (\%) with CS-PIBT after excluding one expert while retaining router-training data from all four distributions.
Parentheses give changes in percentage points relative to the full four-expert MoE.
}
\label{tab:expert-exclusion}
\small
\setlength{\tabcolsep}{3pt}
\renewcommand{\arraystretch}{1.15}
\begin{tabular*}{\linewidth}{@{\extracolsep{\fill}}llrrrrrrrr@{}}
\toprule
& Variant & \mapname{Maze} & \mapname{WH.} & \mapname{Room} & \mapname{Random} & \mapname{Sort.} & \mapname{Stripes} & \mapname{Empty} & Avg. \\
\midrule
\multirow{5}{*}{\rotatebox{90}{\footnotesize MAGAT-MoE}}
& $\setminus$ Maze & 75.0 (+0.2) & 89.2 (-2.0) & 88.2 (-1.4) & \textbf{87.4} (+3.0) & \textbf{99.0} (+0.0) & 45.3 (-1.0) & 100.0 (+0.0) & 83.4 (-0.2) \\
& $\setminus$ WH. & 77.0 (+2.2) & 87.6 (-3.6) & 87.6 (-2.0) & 87.2 (+2.8) & \textbf{99.0} (+0.0) & 43.7 (-2.7) & 100.0 (+0.0) & 83.2 (-0.5) \\
& $\setminus$ Room & 74.8 (+0.0) & 89.4 (-1.8) & 83.8 (-5.8) & 87.0 (+2.6) & 98.8 (-0.2) & 43.7 (-2.7) & 100.0 (+0.0) & 82.5 (-1.1) \\
& $\setminus$ Random & \textbf{77.2} (+2.4) & 89.8 (-1.4) & 88.8 (-0.8) & 84.4 (+0.0) & 98.6 (-0.4) & \textbf{46.7} (+0.3) & 100.0 (+0.0) & \textbf{83.6} (+0.0) \\
\cmidrule(l){2-10}
& Full MoE & 74.8 & \textbf{91.2} & \textbf{89.6} & 84.4 & \textbf{99.0} & 46.3 & 100.0 & \textbf{83.6} \\
\midrule
\multirow{5}{*}{\rotatebox{90}{\footnotesize \modelname{}-MoE}}
& $\setminus$ Maze & 85.4 (-1.6) & 95.4 (-0.2) & 96.4 (-1.0) & 93.0 (+0.4) & 100.0 (+0.2) & 66.0 (-1.3) & \textbf{100.0} (+0.0) & 90.9 (-0.5) \\
& $\setminus$ WH. & 86.0 (-1.0) & 90.8 (-4.8) & 97.6 (+0.2) & 91.4 (-1.2) & 100.0 (+0.2) & \textbf{73.3} (+6.0) & \textbf{100.0} (+0.0) & 91.3 (-0.1) \\
& $\setminus$ Room & \textbf{88.0} (+1.0) & \textbf{96.2} (+0.6) & 95.4 (-2.0) & \textbf{93.2} (+0.6) & 99.8 (+0.0) & 70.3 (+3.0) & 100.0 (+0.0) & \textbf{91.8} (+0.5) \\
& $\setminus$ Random & 87.2 (+0.2) & 96.0 (+0.4) & 97.8 (+0.4) & 89.6 (-3.0) & 99.8 (+0.0) & 71.3 (+4.0) & 100.0 (+0.0) & 91.7 (+0.3) \\
\cmidrule(l){2-10}
& Full MoE & 87.0 & 95.6 & 97.4 & 92.6 & 99.8 & 67.3 & 100.0 & 91.4 \\
\bottomrule
\end{tabular*}%
\end{table}
}

\end{document}